\documentclass[sigconf,natbib,authorversion,nonacm]{acmart}
\usepackage{algorithm}
\usepackage{algorithmic}
\usepackage[utf8]{inputenc}
\usepackage{pgfplots}
\DeclareUnicodeCharacter{2212}{−}
\usepgfplotslibrary{groupplots,dateplot}
\usetikzlibrary{patterns,shapes.arrows}
\pgfplotsset{compat=newest}
\usepackage{adjustbox}
\usepackage{bbm}
\usepackage{enumitem}
\AtBeginDocument{%
  \providecommand\BibTeX{{%
    \normalfont B\kern-0.5em{\scshape i\kern-0.25em b}\kern-0.8em\TeX}}}

\setcopyright{acmcopyright}
\copyrightyear{2023}
\acmYear{2023}
\acmDOI{XXXXXXX.XXXXXXX}

\usepackage{todonotes}

\acmConference[SIGIR '23]{the 46th International ACM SIGIR Conference on Research and Development in Information Retrieval}{July 23--27,
  2023}{Taipei, Taiwan}
\begin{document}

\title{The Challenge of Differentially Private Screening Rules}

\author{Amol Khanna}
\email{Khanna_Amol@bah.com}
\orcid{0000-0002-5566-095X}
\affiliation{%
  \institution{Booz Allen Hamilton}
  \city{Annapolis Junction}
  \state{MD}
  \country{USA}
}

\author{Fred Lu}
\email{Lu_Fred@bah.com}
\affiliation{%
  \institution{Booz Allen Hamilton}
  \city{Annapolis Junction}
  \state{MD}
  \country{USA}
  \institution{University of Maryland}
  \city{Baltimore County}
  \state{MD}
  \country{USA}
}

\author{Edward Raff}
\email{Raff_Edward@bah.com}
\affiliation{%
  \institution{Booz Allen Hamilton}
  \city{Annapolis Junction}
  \state{MD}
  \country{USA}
  \institution{University of Maryland}
  \city{Baltimore County}
  \state{MD}
  \country{USA}
}

\begin{abstract}
Linear $L_1$-regularized models have remained one of the simplest and most effective tools in data analysis, especially in information retrieval problems where n-grams over text with TF-IDF or Okapi feature values are a strong and easy baseline. 
Over the past decade, screening rules have risen in popularity as a way to reduce the runtime for producing the sparse regression weights of $L_1$ models. However, despite the increasing need of privacy-preserving models in information retrieval, to the best of our knoweledge, no differentially private screening rule exists. In this paper, we develop the first differentially private screening rule for linear and logistic regression. In doing so, we discover difficulties in the task of making a useful private screening rule due to the amount of noise added to ensure privacy. We provide theoretical arguments and experimental evidence that this difficulty arises from the screening step itself and not the private optimizer. Based on our results, we highlight that developing an effective private $L_1$ screening method is an open problem in the differential privacy literature. 
\end{abstract}

\ccsdesc[300]{Computing methodologies~Machine learning}
\ccsdesc[300]{Security and privacy~Formal methods and theory of security}

\keywords{Differential Privacy, Sparse, Regression, Screening}

\received{21 February 2023}
\received[revised]{12 March 2009}
\received[accepted]{5 June 2009}

\maketitle

\section{Introduction}

Screening rules are methods which discard features that do not contribute to a statistical model during training. They are most often used in $L_1$-regularized or $L_1$-constrained linear and logistic regression to set the coefficients of unimportant features to zero during the optimization process. In doing so, they improve the generalization of a model by counteracting overfitting and enable faster convergence. Screening rules have been used to improve the performance of sparse regression optimization on numerous datasets over the past decade and are even included the popular \texttt{R} package \texttt{glmnet} \cite{wang2013lasso, wang2014safe, raj2016screening, ghaoui2010safe, olbrich2015screening, tibshirani2012strong, friedman2021package}. However, to the best of our knowledge, no differentially private screening rule exists. 

Differential privacy is a statistical technique which provides a guarantee of the privacy of training data when building a model. Specifically, given privacy parameters $\epsilon$ and $\delta$ and any two datasets $\mathcal{D}$ and $\mathcal{D}'$ differing on one datapoint, an approximately differentially private algorithm $\mathcal{A}$ satisfies $\mathbb{P} \left[ \mathcal{A}(\mathcal{D}) \in O \right] \leq \\ \exp\{ \epsilon\} \mathbb{P} \left[ \mathcal{A}(\mathcal{D}') \in O \right] + \delta$ for any $O \subseteq \text{image}(\mathcal{A})$ \cite{dwork2014algorithmic}. Intuitively, this means that the output of $\mathcal{A}$ does not reveal whether or not a specific datapoint was used in training. Differential privacy is currently the most effective method to build statistical models on sensitive data (eg. medical records) for publication without betraying the privacy of individuals in the training data and could support private information retrieval \cite{khanna2022privacy,Toledo2016LowerCostI,10.1145/3159652.3162006}. 

Currently, there is an interest in sparse differentially private regression algorithms. Private $L_1$-regularized or constrained optimizers have been developed. However, due to the addition of noise, these algorithms are either unable to maintain the sparsity of the output solution or unable to choose a good support set for the solution \cite{kifer2012private, talwar2015nearly, wang2020differential}. Other works attempt to perform private model selection, but they are computationally inefficient and assume an exact level of sparsity of the final solution in order to choose a support set prior to training \cite{lei2018differentially, thakurta2013differentially}. In contrast, nonprivate screening rules do not require a predetermined support set or level of sparsity. They efficiently check a mathematical condition to determine if a feature should be screened to 0, and are implemented with sparse optimizers during training to improve the rate of learning and stability. 

A differentially private screening rule has the potential to combat overfitting and help private optimizers focus on a model's most important features. However, to the best of our knowledge, no differentially private screening rule exists. In this work, we develop and explore a differentially private screening rule, and we show that it is difficult to accurately screen features. We provide an analysis of the private screening rule's behavior to explain the challenge of creating an effective differentially private screening rule. 

\section{Related Works}

Methods to produce private sparse regression weights all suffer in performance due to the addition of noise. Private $L_1$ optimizers must add higher levels of noise when run for more iterations, incentivizing practitioners to run fewer iterations \cite{talwar2015nearly, wang2020differential}. However, running an optimizer for fewer iterations means that the model will be limited in its learning. On the other hand, private model selection algorithms are computationally inefficient and run prior to training, meaning they are unable to reap the benefit of any information contained within partially trained coefficients of the weight vector \cite{lei2018differentially, thakurta2013differentially}. Although noise is necessary for privacy, an effective private screening rule would run with a private optimizer and improve the optimizer's performance by setting the coefficients of irrelevant features to 0. By using the screening rule on the current weight vector, it can adapt to the optimizer's updates and screen features more accurately.

To develop a differentially private screening rule, we adapt Raj et al.'s rule which is flexible to solving many types of regression problems \cite{raj2016screening}. While other screening rules exist, they are geometry- and problem-specific \cite{ghaoui2010safe, wang2014safe, wang2013lasso}. We utilize their screening rule for $L_1$-constrained regression to determine whether it can improve the sparsity of Talwar et al.'s $L_1$-constrained private Frank-Wolfe algorithm \cite{raj2016screening, talwar2015nearly}. To the best of our knowledge, this is the first work considering a differentially private screening rule. Although our experiments find that the noise added to the screening rule overpowers its ability to screen regression weights correctly, we provide an analysis of this result to shed insight into the challenge of creating an effective private screening rule. We believe this work is valuable as it explores a novel approach to private sparsity and provides a baseline for the development and implementation of differentially private screening rules. 

\section{Methods}
\label{sec:3}

Raj et al. consider the problem $\min_{\mathbf{w} \in \mathcal{C}} f(\mathbf{Xw})$, where $\mathcal{C}$ is the feasible set of solutions and $f$ is $L$-smooth and $\mu$-strongly convex \cite{raj2016screening}. They also define $\mathbf{x}_{(i)}$ to be the $i^{\text{th}}$ column of the design matrix $\mathbf{X}$ and $\mathcal{G}_{\mathcal{C}}(\mathbf{w})$ to be the Wolfe gap function, namely $\max_{\mathbf{z} \in \mathcal{C}} (\mathbf{Xw} - \mathbf{Xz})^{\top}\nabla f(\mathbf{Xw})$. Given this information, they prove that if
\begin{align}
\label{eq:1}
    &\lvert \mathbf{x}_{(i)}^{\top} \nabla f(\mathbf{Xw}) \rvert \nonumber  + (\mathbf{Xw})^{\top}\nabla f(\mathbf{Xw}) \\ + \ &L(\lVert \mathbf{x}_{(i)} \rVert_2 + \lVert \mathbf{Xw} \rVert_2) \sqrt{\mathcal{G}_{\mathcal{C}}(\mathbf{w}) / \mu}
\end{align}
is less than $0$, then $w_{i}^{*} = 0$, where $\mathbf{w}^{*}$ is the optimal solution to the optimization problem. Our goal is to determine the sensitivity of this calculation so we can add an appropriate amount of noise and ensure screening is differentially private. 

We will conduct our analysis for the case where $\lVert\mathbf{x}_i \rVert_\infty \leq 1$ and $\mathcal{C}$ is the $\lambda$-scaled $L_1$-ball in $\mathbb{R}^{d}$. These conditions are also required by Talwar et al.'s private Frank-Wolfe algorithm, which we use for $L_1$-constrained optimization in \hyperref[sec:implementing-private-screening]{Section 3.3} \cite{talwar2015nearly}. 

\subsection{Sensitivity of Linear Regression}
\label{sec:3.1}

Let $\mathbf{u} = \mathbf{Xw}$. For linear regression, $f(\mathbf{u}) = \frac{1}{2n} (\mathbf{u} - \mathbf{y})^{\top}(\mathbf{u} - \mathbf{y})$, implying $\nabla f(\mathbf{u}) = \frac{1}{n}(\mathbf{u} - \mathbf{y})$ and $\nabla^{2} f(\mathbf{u}) = \frac{1}{n} \mathbf{I}_{n}$. Therefore, from the definitions of Lipschitz smoothness and strong convexity, we can see that $f(\mathbf{u})$ is $\frac{1}{n}$-smooth and $\frac{1}{n}$-strongly convex with respect to $\mathbf{u}$. 

By using the triangle inequality and the fact that the maximum of a sum is at most the sum of each element's maximum, we can bound the sensitivity of \hyperref[eq:1]{Equation 1} for linear regression by summing the sensitivity of each of its terms. These calculations are shown below. Assume without loss of generality that $\mathbf{X}$ and $\mathbf{X}'$ differ in their first row for ease of notation. 

To bound the first term of \hyperref[eq:1]{Equation 1}, note that 
\begin{align*}
    &\max_{\mathbf{X}, \mathbf{X}'} \  \left\lvert \left\langle \mathbf{x}_{(i)}, \nabla f(\mathbf{Xw}) \right\rangle - \left\langle \mathbf{x}_{(i)}', \nabla f(\mathbf{X}'\mathbf{w}) \right\rangle \right\rvert \\ 
    = & \max_{\mathbf{X}, \mathbf{X}'} \ \left\lvert x_{1i} \left[\frac{1}{n}\left(\mathbf{x}_1^{\top}\mathbf{w} - y_1 \right)  \right] - x_{1i}' \left[\frac{1}{n}\left({\mathbf{x}'}_1^{\top}\mathbf{w} - y_1 \right)  \right] \right\rvert \\ 
    \leq & \frac{2\lambda}{n},
\end{align*}
where the first simplification comes from expanding the inner products and the second follows directly from the triangle inequality and restrictions on the feasible set and norms of the input data. This means that the sensitivity of the first term for one value of $i$ is $\frac{2\lambda}{n}$. By the sensitivity principles of vectors, this means that releasing the values of the first term for all $i$ has $L_2$-sensitivity of $\frac{2\lambda\sqrt{d}}{n}$ \cite{dwork2014algorithmic}. 

For the second term, 
\begin{align*}
    &\max_{\mathbf{X}, \mathbf{X}'} \  \lvert \langle \mathbf{Xw}, \nabla f(\mathbf{Xw}) \rangle - \langle \mathbf{X}'\mathbf{w}, \nabla f(\mathbf{X}'\mathbf{w}) \rangle \rvert \\ = & \max_{\mathbf{X}, \mathbf{X}'} \ \left\lvert \mathbf{x}_{1}^{\top}\mathbf{w} \left[\frac{1}{n}\left(\mathbf{x}_1^{\top}\mathbf{w} - y_1 \right)  \right] - {\mathbf{x}'}_1^{\top}\mathbf{w} \left[\frac{1}{n}\left({\mathbf{x}'}_1^{\top}\mathbf{w} - y_1 \right)  \right] \right\rvert \\ \leq &\frac{2\lambda^2}{n},
\end{align*}
using the same logic as the previous calculation. This means its sensitivity is $\frac{2\lambda^2}{n}$ and its $L_2$-sensitivity is $\frac{2\lambda^2 \sqrt{d}}{n}$.

To find the sensitivity of the third term, we note that the maximum of a product is bounded by the product of maximums. Given this, we find that 
\begin{align*}
    &\max_{\mathbf{X}, \mathbf{X}'} \  \left\lvert \left\lVert \mathbf{x}_{(i)} \right\rVert_2 - \left\lVert \mathbf{x}'_{(i)} \right\rVert_2 \right\rvert 
    \leq \max_{\mathbf{X}, \mathbf{X}'} \sqrt{\left\lvert x_{1i}^2 - {x'}_{1i}^2 \right\rvert}  
    \leq 1,
\end{align*}
where the first simplification is derived from expanding the first and noting that the difference of square roots must be less than or equal to the square root of the absolute value of the difference in their squared terms. The same logic can be applied for 
\begin{align*}
    &\max_{\mathbf{X}, \mathbf{X}'} \  \left\lvert \left\lVert \mathbf{Xw} \right\rVert_2 - \left\lVert \mathbf{X}'\mathbf{w} \right\rVert_2 \right\rvert 
    \leq  \max_{\mathbf{X}, \mathbf{X}'} \  \sqrt{\left\lvert \left(\mathbf{x}_1^{\top}\mathbf{w}\right)^2 - \left({\mathbf{x}'}_1^{\top}\mathbf{w}\right)^2 \right\rvert}  
    \leq \lambda.
\end{align*}
For the Wolfe gap function, 
\begin{align*}
    &\max_{\mathbf{X}, \mathbf{X}', \mathbf{z} \in \mathcal{C}} \ \lvert \langle \mathbf{Xw} - \mathbf{Xz}, \nabla f(\mathbf{Xw}) \rangle - \langle \mathbf{X'w} - \mathbf{X'z}, \nabla f(\mathbf{X'w}) \rangle \rvert \\ 
    = &\max_{\mathbf{X}, \mathbf{X}', \mathbf{z} \in \mathcal{C}} 
        \left\lvert \mathbf{x}_{1}^{\top}(\mathbf{w} - \mathbf{z}) \left[\frac{1}{n}\left(\mathbf{x}_1^{\top}\mathbf{w} - y_1 \right)  \right] \right. \\  
    & \left. \mathrel{\phantom{\max_{\mathbf{X}, \mathbf{X}', \mathbf{z} \in \mathcal{C}}}} - {\mathbf{x}'}_1^{\top}(\mathbf{w} - \mathbf{z}) \left[\frac{1}{n}\left({\mathbf{x}'}_1^{\top}\mathbf{w} - y_1 \right)  \right] \right\rvert \\
    \leq &\frac{4 \lambda^2}{n}. 
\end{align*}
Plugging in each of these calculations, the $L_2$-sensitivity of the third term is bounded by $\frac{1}{n} \left( \sqrt{d} + \lambda\sqrt{d} \right)\sqrt{\frac{4\lambda^2/n}{1/n}}$. This means the total $L_2$-sensitivity of the screening rule is bounded by 
\begin{equation}
\label{eq:2}
\frac{2\lambda\sqrt{d}}{n} + \frac{2\lambda^2\sqrt{d}}{n} + \frac{1}{n} \left( \sqrt{d} + \lambda\sqrt{d} \right)\sqrt{\frac{4\lambda^2/n}{1/n}}.
\end{equation}

\subsection{Sensitivity of Logistic Regression}
For logistic regression, the binary cross-entropy loss is $f(\mathbf{u}) = -\frac{1}{n} \sum y_i \log \sigma(u_i) + (1 - y_i) \log (1 - \sigma(u_i))$, where $\sigma$ is the element-wise sigmoid function. This implies $\nabla f(\mathbf{u}) = -\frac{1}{n}(\mathbf{y} - \sigma(\mathbf{u}))$ and $\nabla^2 f(\mathbf{u}) = \frac{1}{n} (\sigma(\mathbf{u}) \odot (\mathbf{1} - \sigma(\mathbf{u})))\mathbf{I}_n$, where $\odot$ represents elementwise multiplication. From these equations we find that the binary cross-entropy loss is $\frac{1}{4n}$-smooth and $\frac{\sigma(\lambda)(1 - \sigma(\lambda))}{n}$-strongly convex. 

The derivation for the sensitivity of \hyperref[eq:1]{Equation 1} for logistic regression follows the same steps as used for linear regression in \hyperref[sec:3.1]{Section 3.1}, so we do not show the full calculations here. The final $L_2$-sensitivity bound derived is 
\begin{equation}
\label{eq:3}
    \frac{2\sigma(\lambda)\sqrt{d}}{n} + \frac{2\lambda\sigma(\lambda)\sqrt{d}}{n} + \frac{1}{4n}\left( \sqrt{d} + \lambda\sqrt{d} \right) \sqrt{\frac{4\lambda\sigma(\lambda) / n}{\sigma(\lambda)(1 - \sigma(\lambda)) / n}}.
\end{equation}

\subsection{Implementing Private Screening}

\begin{algorithm}[t]
\label{alg:1}
\caption{$L_1$-Constrained Regression with Screening}
\begin{algorithmic}[1]
\REQUIRE Privacy Parameters: $\epsilon_1 > 0$, $\epsilon_2 > 0$, $0 < \delta_1 \leq 1$, $0 < \delta_2 \leq 1$; Constraint: $\lambda > 0$; Iterations: $T$; Design Matrix: $\mathbf{X} \in \mathbb{R}^{n \times d}$ where $\lVert \mathbf{x}_i \rVert_\infty \leq 1$ for all $i \in \{1, \ldots, n \}$; Target: $\mathbf{y}$; $L_2$-Sensitivity: $s$; Iterations to Screen: $\mathbf{i}$.
\STATE $l \gets \texttt{Length}(\mathbf{i})$
\STATE $\delta_{\text{iter}} \gets \frac{\delta_2}{l + 1}$
\STATE $\epsilon_{\text{iter}} \gets \frac{\epsilon_2}{2\sqrt{2l \log \left(1 / \delta_{\text{iter}}\right)}}$
\STATE $\sigma^2 \gets \frac{2s^2 \log \left( 1.25/\delta_{\text{iter}} \right)}{\epsilon_{\text{iter}}^2}$
\STATE $\widehat{\mathbf{w}}^{(0)} \gets$ Random Vector in the $\lambda$-scaled $L_1$ Ball
\FOR{$t = 1$ to $T$}
    \STATE $\widehat{\mathbf{w}}^{(t)} \gets \texttt{DP-Frank-Wolfe Step} \left( \epsilon_1, \delta_1, \lambda, T, \mathbf{X}, \mathbf{y}, \widehat{\mathbf{w}}^{(t - 1)} \right)$
    \IF{$t \in \mathbf{i}$}
        \STATE $screen \gets \hyperref[eq:1]{\texttt{Equation 1}}\left( \mathbf{X}, \mathbf{y}, \widehat{\mathbf{w}}^{(t)}, \lambda \right)$
        \STATE $screen \gets screen + \mathcal{N}(\mathbf{0}, \sigma^2\mathbf{I}_d)$
        \STATE $\widehat{\mathbf{w}}_j^{(t)} \gets 0$ \textbf{if} $screen_j < 0$ \textbf{for all} $j$
    \ENDIF
\ENDFOR
\STATE Output $\widehat{\mathbf{w}}^{(T)}$
\end{algorithmic}
\end{algorithm}

\label{sec:implementing-private-screening}
Now that we have calculated the bounds for the $L_2$-sensitivity of the screening rule in \hyperref[eq:1]{Equation 1}, we discuss how we implemented it into a differentially private regression training procedure. 

Since our screening rule requires $L_1$-constrained optimization, we employ the private Frank-Wolfe algorithm developed by Talwar et al. to train regression models \cite{talwar2015nearly}. To the best of our knowledge, this is the only differentially-private algorithm for $L_1$-constrained optimization. Additionally, both linear and logistic loss have Lipschitz constant $1$ with respect to the $L_1$ norm, which satisfies the algorithm's requirement for $L_1$-Lipschitz loss functions. 

Our algorithm is shown in \hyperref[alg:1]{Algorithm 1}, abstracting away the steps required for the private Frank-Wolfe algorithm. By using the Gaussian mechanism and the advanced composition theorem for approximate differential privacy, the screening technique in \hyperref[alg:1]{Algorithm 1} is $(\epsilon_2, \delta_2)$-differentially private \cite{dwork2006our, dwork2010boosting}. Following this, the basic composition theorem of approximate differential privacy guarantees that the results of \hyperref[alg:1]{Algorithm 1} is $(\epsilon_1 + \epsilon_2, \delta_1 + \delta_2)$-differentially private \cite{dwork2006our}.

\section{Experiments}
\label{sec:4}

\begin{figure}[t]
    \label{fig:1}
    \centering
    \includegraphics[width=\linewidth]{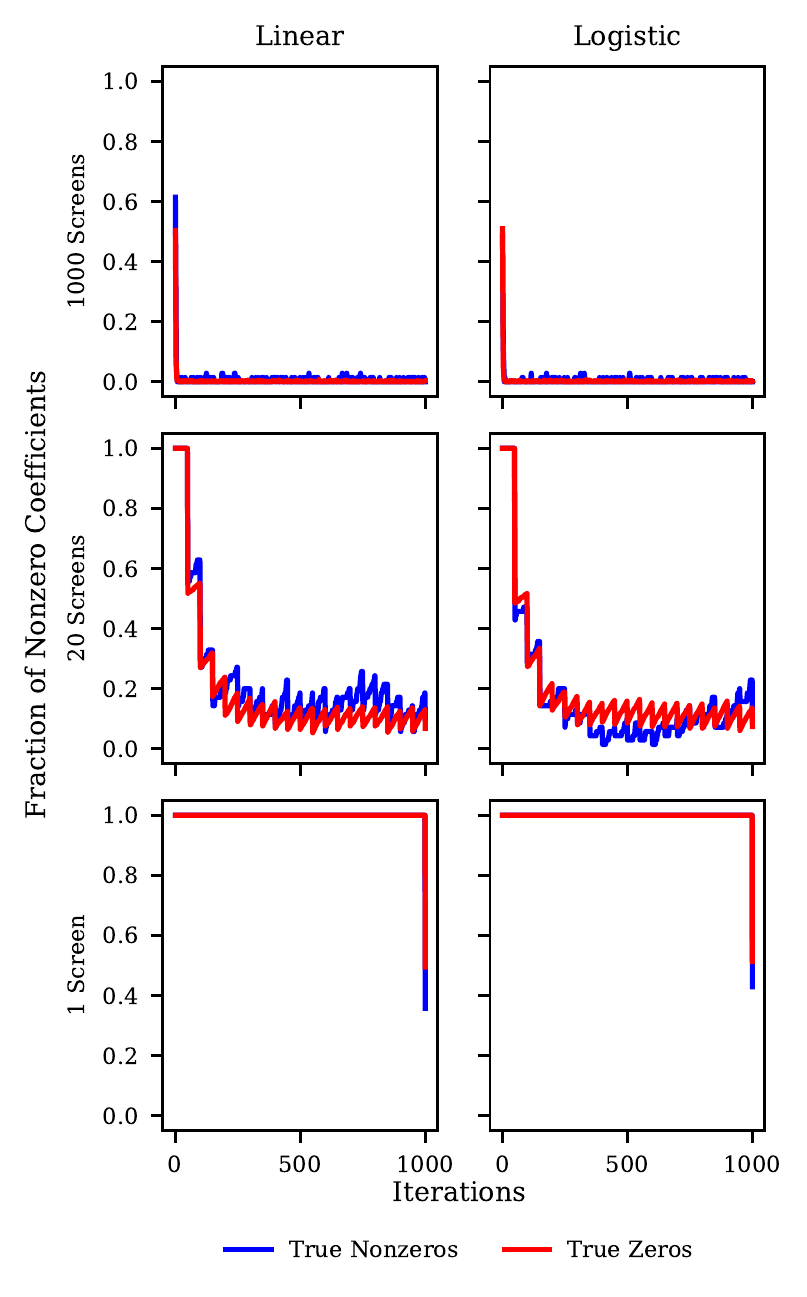}
    \vspace{-0.3 in}
    \caption{Testing \hyperref[alg:1]{Algorithm 1} on a synthetic dataset.}
    \vspace{-0.15 in}
    \Description{The figure displays six subplots for the three cases described in the main body of the paper for linear and logistic regression. None of plots demonstrate that the screening algorithm is able to retain the true nonzero weights significantly better than the true zero weights.}
\end{figure}

To test whether \hyperref[alg:1]{Algorithm 1} performs well in practice, we tested how it performed on linear and logistic regression. To do so, we used the synthetic dataset which Raj et al. used to test their nonprivate screening algorithm \cite{raj2016screening}. Specifically, we generated $3000$ datapoints in $\mathbb{R}^{600}$ from the standard normal distribution, and scaled the final dataset so $\lVert \mathbf{x}_i \rVert_\infty \leq 1$. We set the true weight vector $\mathbf{w}^*$ to be sparse with 35 entries of $+1$ and 35 entries of $-1$. For linear regression, $\mathbf{y} = \mathbf{Xw}^*$, and for logistic regression, $\mathbf{y} = \mathbbm{1}_{\mathbf{Xw}^* > 0}$. Raj et al. demonstrated that the nonprivate screening rule listed in \hyperref[eq:1]{Equation 1} performs well on this dataset for linear regression \cite{raj2016screening}. We verified this result, finding that using the nonprivate Frank-Wolfe optimizer with the nonprivate screening rule at every iteration produced a final weight vector in which nonzero components were only at the locations of nonzero components in the true weight vector and $55\%$ of the true nonzero components were nonzero after training in both linear and logistic regression. We then ran the private \hyperref[alg:1]{Algorithm 1} on this dataset for 1000 iterations with $\epsilon_1 = \epsilon_2 = 2.5, \delta_1 = \delta_2 = \frac{1}{6000}, \text{ and }\lambda = 5$.

\begin{figure}[t]
    \label{fig:2}
    \centering
    \includegraphics[width=\linewidth]{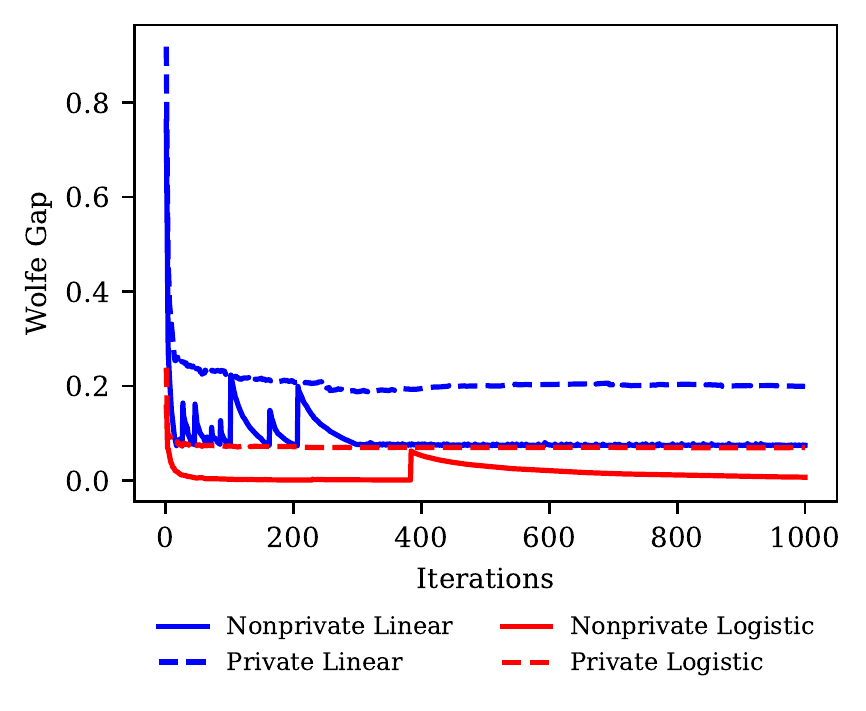}
    \vspace{-0.3 in}
    \caption{Comparing the Wolfe gap function for nonprivate and private optimization when nonprivate screening is applied at every iteration.}
    \vspace{-0.25 in}
    \Description{The figure demonstrates that nonprivate optimization produces lower values of the Wolfe gap function compared to private optimization.}
\end{figure}

\hyperref[fig:1]{Figure 1} shows the results of this experiment when we implemented screening after every iteration, every $50^{\text{th}}$ iteration, and after the last iteration. It is clear that \hyperref[alg:1]{Algorithm 1} is not able to discriminate between screening true zero and true nonzero coefficients in any of these cases. Additionally, when private screening is implemented too often, it screens too many coefficients, and since the Frank-Wolfe algorithm only updates one coefficient at a time, after a few iterations, the weight vector is only able to have up to a few nonzero coefficients which have not been screened to zero. To identify whether the private Frank-Wolfe algorithm or the private screening methods were causing these poor results, we ran the following two experiments: 
\begin{enumerate}[label=(\Alph*)]
    \item \label{exp:A} We tested how well nonprivate screening performed using the private Frank-Wolfe algorithm with $\epsilon = 2.5$ and $\delta=\frac{1}{6000}$. We found that no matter how often we implemented the screening rule, no coefficients were screened from the solution. 
    \item \label{exp:B} We tested how well private screening performed using the nonprivate Frank-Wolfe algorithm with $\epsilon=2.5$ and $\delta=\frac{1}{6000}$. We found that when screening every $50^{\text{th}}$ iteration, the screening rule would produce a final weight vector with nonzero components in approximately $10\%$ of the true nonzero components and none of the true zero components. The results of this experiment when screening every iteration or only at the last iteration mimicked those found in the respective rows of \hyperref[fig:1]{Figure 1}.
\end{enumerate}
These experiments provide key insights into the results shown in \hyperref[fig:1]{Figure 1}. \hyperref[exp:A]{Experiment A} suggests that all of the screening occurring in \hyperref[fig:1]{Figure 1} arises from noise added to the screening rule. This is because without the screening's noise, no screening occurs. It also implies that the noise added for private optimization made it more difficult to screen coefficients, since when we tested completely nonprivate screening (without noisy optimization), the nonprivate screening rule worked well. Heuristically, this outcome may arise because noisy weights make the Wolfe gap function in \hyperref[eq:1]{Equation 1} very large, meaning that it overpowers the second term, which is the only term that can be negative and is thus essential to effective screening. We verified that the Wolfe gap function evaluates to smaller values when using nonprivate optimization for both linear and logistic regression. This result can be seen in \hyperref[fig:2]{Figure 2}.

\hyperref[exp:B]{Experiment B} indicates that the noise added in the private screening rule makes it much stronger that its nonprivate counterpart. This is observed by noting that without a noisy screening rule, screening at every iteration with the nonprivate Frank-Wolfe optimizer would not screen all the true nonzero components to zero, whereas with the noisy screening rule, almost all components are screened to zero after only a few iterations. 

\section{Discussion}

The goal of a differentially private screening rule is to improve sparse private optimization during training. The screening rule described in \hyperref[alg:1]{Algorithm 1} is computationally efficient and able to benefit from the information contained in partially trained coefficients, unlike private model selection algorithms in prior works. However, the results in \hyperref[fig:1]{Figure 1} indicate that it is not effective at screening only true zero coefficients. Given the sensitivities derived in \hyperref[eq:2]{Equation 2} and \hyperref[eq:3]{Equation 3} and the results in \hyperref[sec:4]{Section 4}, we use this section to discuss the major challenge to developing an effective differentially private screening rule. 

The sensitivities of linear and logistic regression are on the order of $\sqrt{d}$, so a private screening rule would have to add noise with a scale of $\mathcal{O}\left(\sqrt{d}\right)$. This is a property of private sensitivity, and noise with a scale of $\mathcal{O}\left(\sqrt{d}\right)$ is also found in private optimization and private model selection algorithms. Our results clearly show this noise level is too large. We also note the sample complexity of non-private $L_1$ models is known to grow at a $\mathcal{O}(\log d)$ rate, seemingly implying that our screening model may not work well asymptotically.

Though discouraging, our results leave a number of open questions now that we have identified the difficulty of private screening rules. 1) Can differentailly private screening rules be effective in the finite values of $d$ that occur in practice? 2) Can the noise added by screening be reduced to a rate of $\mathcal{O}(\log d)$? 3) How can the concept of a ``safe'' screening rule be adapted for differential privacy, since no screening rule can avoid false-positives when noise is added? 

\section{Conclusion}

In this paper, we are the first to consider differentially private screening rules. We attempt to develop such a rule by modifying a general-purpose nonprivate screening rule, but when testing our algorithm on a synthetic dataset, we find that noisy optimization and screening produces poor performance. By analyzing how different sources of noise affect the screening rule's behavior, we identify the limitations to our algorithm. We conclude by discussing the challenges to developing a useful private screening rule. We highlight that developing an effective differentially private screening rule is an open problem with the potential to improve the efficiency and accuracy of high dimensional private regression. 

\bibliographystyle{ACM-Reference-Format}
\bibliography{bibliography}

\end{document}